\documentclass[a4paper]{article}

\usepackage{url}
\usepackage{nicefrac}
\usepackage{algorithmic}
\usepackage{algorithm}
\usepackage{amsmath}
\usepackage{amssymb}
\usepackage{natbib}
\usepackage{graphicx}
\usepackage{anysize}
\usepackage[color]{changebar}
\usepackage[affil-it]{authblk}
\usepackage{ntheorem}

\marginsize{1.5cm}{1.5cm}{1cm}{1cm}

\newcommand{\infig}[1]{{\textsf{#1}}}
\newcommand{\Rex}{\ensuremath{R_{\textrm{ex}}}}

\newcommand{\Rpi}{\ensuremath{R_\textrm{in}^{\textrm{PI}}}} \newcommand{\Rh}
{\ensuremath{R_\textrm{in}^{\textrm{H}}}}

\graphicspath{{}}

\author[1]{Keyan Zahedi}
\author[1]{Georg Martius}
\author[1,2]{Nihat Ay}
\affil[1]{Information Theory of Cognitive Systems, Max Planck Institute for
  Mathematics in the Sciences, Leipzig, Saxony, Germany}
\affil[2]{%
Santa Fe Institute, 1399 Hyde Park Road, Santa Fe, New Mexico 87501, USA
}

\title{Linear combination of one-step predictive information with an external reward in an episodic policy gradient setting: a critical analysis}

\begin{document}
\maketitle

\begin{abstract}
  One of the main challenges in the field of embodied artificial intelligence is
  the open-ended autonomous learning of complex behaviours. Our approach is to use
  task-independent, information-driven intrinsic motivation(s) to support
  task-dependent learning. The work presented here is a preliminary step in which
  we investigate the predictive information (the mutual information of the past
  and future of the sensor stream) as an intrinsic drive, ideally supporting any
  kind of task acquisition. Previous experiments have shown that the predictive
  information (PI) is a good candidate to support autonomous, open-ended
  learning of complex behaviours, because a maximisation of the PI corresponds to an
  exploration of morphology- and environment-dependent behavioural regularities.
  The idea is that these regularities can then be exploited in order to solve
  any given task. Three different experiments are presented and their results
  lead to the conclusion that the linear combination of the one-step PI with an
  external reward function is not generally recommended in an episodic policy gradient
  setting. Only for hard tasks a great speed-up can be achieved at the cost of an
  asymptotic performance lost.
  \medskip\\

\end{abstract}

\section{Introduction}
One of the main challenges in the field of embodied artificial intelligence (EAI)
is the open-ended autonomous learning of complex behaviours. Our approach is to
use task-independent, information-driven intrinsic motivation to support
task-dependent learning in the context of reinforcement learning (RL) and EAI.
The work presented here is a first step into this direction. RL is of growing
importance in the field of EAI, mainly for two reasons. First, it allows to
learn the behaviours of high-dimensional and complex systems with simple
objective functions. Second, it has a well-established theoretical
\citep{Sutton1998Reinforcement-Learning:-An,Bellman2003Dynamic-Programming} and
biological foundation \citep{Dayan2002Reward-Motivation-and-Reinforcement}. In
the context of EAI, where the agent has a morphology and is situated in an
environment, the concepts of the agent's intrinsic and extrinsic perspective
rise naturally. As a direct consequence, several questions about intrinsic and
extrinsic reward functions, denoted by IRF and ERF, follow from the EAI's point
of view. The questions that are of interest to us are; what distinguishes an IRF
from an ERF, what is a good candidate for a first principled IRF, and finally,
how should IRFs and ERFs be combined?

The first question, of how to distinguish between IRF and ERF is addressed in
the second section of this work, which starts with the conceptual framework of
the sensorimotor loop and its representation as a causal
graph. This leads to a natural distinction of variables that are intrinsic and
extrinsic to the agent. We define an IRF that models an internal drive or
motivation as a task-independent function which operates on the agent's
intrinsic variables only. In general, an ERF is a task-dependent function that may operate
on intrinsic and extrinsic variables.

The main focus of this work is the second question, which deals with finding a
first principled IRF. We propose the predictive
information (PI) \citep{Bialek1999Predictive-Information} for the following
reasons. Information-driven self-organisation, by the means of maximising the
one-step approximation of the PI has proved to produce a coordinated behaviour
among physically coupled but otherwise independent agents
\citep{Zahedi2010Higher-coordination-with,Ay2008Predictive-information-and}. The
reason is that the PI inherently addresses two important issues of
self-organised adaptation, as the following equation shows: $I(S_t;S_{t+1}) =
H(S_{t+1}) - H(S_{t+1}|S_t)$, where $S_t$ is the vector of intrinsically
accessible sensor values at time $t$. The first term leads to a diversity of the behaviour,
as every possible sensor state must be visited with equal probability. The
second term ensures that the behaviour is compliant with the constraints given
by the environment and the morphology, as the behaviour must be predictable. This
means that an agent maximising the PI explores behavioural regularities,
which can then be exploited to solve a task.
In a differently motivated work, namely to obtain purely self-organising behaviour,
a time-local version of the PI was successfully used to drive the learning process
of a robot controller \citep{MartiusDerAy2013}. A similar learning rule was obtained
from the principle of Homeokinesis \citep{Der2012The-Playful-Machine:-Theoretical}.
In both cases a gradient information was derived to pursue local optimisation.
For the integration of external goals a set of methods have been proposed~\cite{martiusherrmann:variantsofgso12}, which however cannot deal with the standard reinforcement setting of
 arbitrary time-delayed rewards that we study here.
\citet{Prokopenko2006Evolving-Spatiotemporal-Coordination}
used the PI, estimated on the
spatio-temporal phase-space of an embodied system, as part of fitness function
in an artificial evolution setting.
 It was shown that
the resulting locomotion behaviour of a snake-bot was more robust, compared to
the setting, in which only the travelled distance determined the fitness.

The third question, which deals with how to combine the IRF and ERF, is in the
focus of the ongoing research that was briefly described above and of which this
publication is a first step. As the PI maximisation is considered to be an
exploration of behavioural regularities, it would be natural to exchange the
exploration method of a RL algorithm by a gradient on the PI. The work presented
here is a preliminary step in which we concentrate on the effect of the PI in a RL
context to understand for which type of learning problems it is beneficial and in
which it might not be. Therefore, we chose a linear combination of IRF and ERF
in an episodic RL setting to evaluate the PI as an IRF in different experiments.
Combining an IRF and an ERF in this way is justified as ERFs
are often linear combinations of different terms, such as one term for
fast locomotion and another for low energy consumption. Nevertheless, the results of
the experiments presented in this work show that the one-step PI should not be
combined in this way with an ERF in an episodic policy gradient setting.

We are not the first to address the question of IRF and ERF in the context of RL
and EAI. This idea goes back to the pioneering work
of~\citet{Schmidhuber1990A-possibility-for} and is also in the focus of more
recent work by \citet{Kaplan2004Maximizing-Learning-Progress:,Schmidhuber:06cs,%
Oudeyer2007Intrinsic-Motivation-Systems} based on prediction progress or
prediction error \citep{Barto2004Intrinsically-motivated-learning}. In
\citet{Storck:95,sunyi2011agi} an intrinsic reward for information gain was
proposed (KL-divergence between subsequent models), which results in their
experiments in a state-entropy maximisation. A different approach
\citep{Little2013Learning-and-exploration-in-action-perception} uses a greedy
policy on the expected information gain of the world model to select the next
action of an agent. However only discrete state/action spaces have been
considered in both approaches. A similar work \citep{cuccu2011} uses compression
quality as the intrinsic motivation, which was particularly beneficial because
it performed a reduction of the high-dimensional visual input space. In
comparison to our work only one experiment (similarly structured than the
self-rescue task below) with an one-dimensional action-space was used without
considering asymptotic performance, where we found most problems.

This paper investigates continuous space high-dimensional control problems
 where random exploration becomes difficult.
The PI, measured on the sensor values, accompanies (and might eventually
replace) the exploration of a RL method such that the policy adaptations are
conducted compliant to the morphology and environment. The actual embodiment is
 taken into account, without modelling it explicitly in the learning process.

The work is organised as in the following way. The next section gives an
overview of the methods, beginning with the sensorimotor loop
and its causal representation. This is then followed by a presentation of the PI and the
episodic RL method PGPE
\citep{Sehnke2010Parameter-exploring-policy-gradients}. The third section
describes the results received by applying the methods to three experiments, and
the last section closes with a discussion.

\section{Methods}
This section describes the methods used in this work. It begins with the
conceptual framework of the sensorimotor
loop. This is then followed by a discussion of the PI and
entropy, which both are used as IRF in all presented
experiments. Finally, the RL algorithm utilised in this
work is introduced as far as it is required to understand how the results were
obtained.

\subsection{Embodied Agents and the Sensorimotor Loop}
\label{sec:sml}
\begin{figure}
  \begin{center}
    \includegraphics[width=\textwidth]{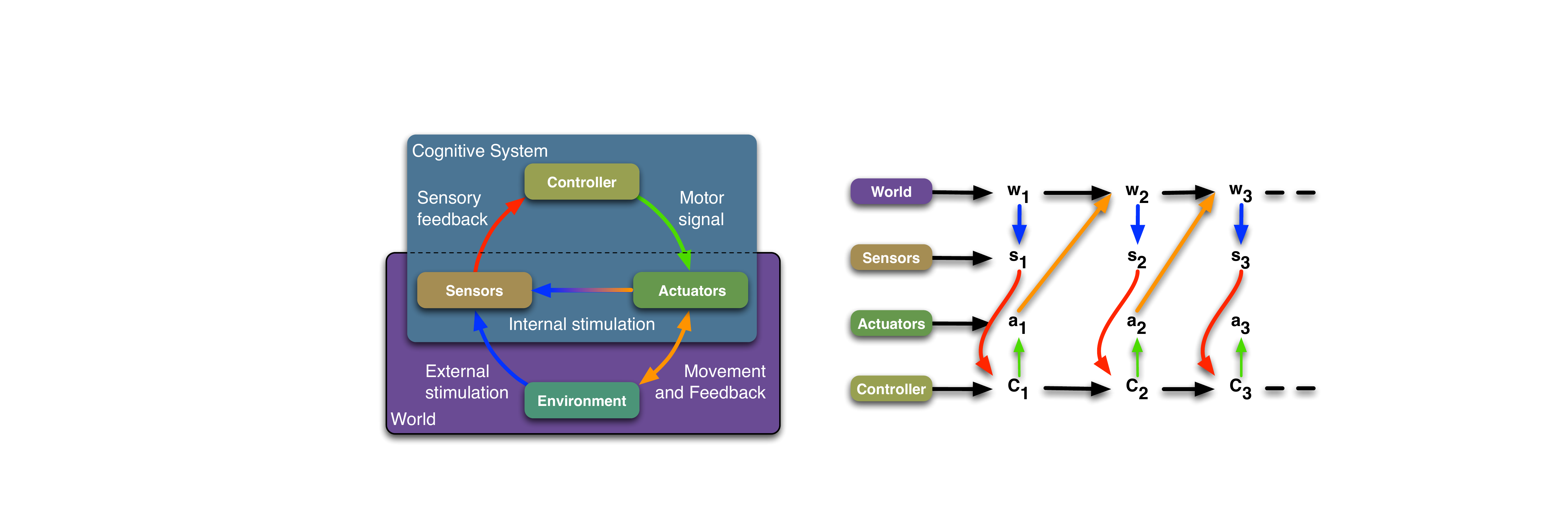}
  \end{center}
  \caption{The sensorimotor loop. Left: schematic diagram of a cognitive system
    with its interaction with the world. Right: Corresponding causal graph.}
  \label{fig:sml}
\end{figure}

There are three main reasons why we prefer to experiment with embodied agents
(EA). First, \emph{scalability}: EA are high-dimensional systems
which live in a continuous world. Hence, the algorithms face the curse
of dimensionality if they are
evaluated on different EAs. Second, \emph{validation}: we are interested in
understanding natural cognitive systems by the means of building artificial agents
\citep{Brooks1991Intelligence-Without-Reason}. Using EA ensures that the
models are validated against the same (or similar) physical constraints that
natural systems are exposed to. Third, \emph{guidance:} there is good
evidence that the constraints posed by the morphology and environment can be
used to reduce the required controller complexity, and hence, reduce the
size of the search space for a learning algorithm
\citep{Zahedi2010Higher-coordination-with,Pfeifer2006How-the-Body}. Consequently, understanding the
interplay between the body, brain and environment, also called the sensorimotor
loop (SML, see Fig.~\ref{fig:sml}), is a general focus of our work. The next
paragraph will introduce the general concept of the SML and discuss its
representation as a causal graph.

A cognitive system consists of a brain or controller that sends signals to the
system's actuators, which then affect the system's environment. We prefer the
notion of the system's \emph{Umwelt}
\citep{Uexkuell1957A-Stroll-Through,Clark1996Being-There:-Putting,Zahedi2010Higher-coordination-with,Zahedi2013Quantifying-Morphological-Computation},
which is the part of the system's environment that can be affected by the
system, and which itself affects the system. The state of the actuators and the
\emph{Umwelt} are not directly accessible to the cognitive system, but the loop
is closed as information about both, the \emph{Umwelt} and the actuators are
provided to the controller by the system's sensors. In addition to this general
concept, which is widely used in the EAI community
\citep[see e.g.][]{Pfeifer2007Self-Organization-Embodiment-and}, we introduce the
notion of \emph{world} to the sensorimotor loop, and by that we mean the
system's morphology and the system's \emph{Umwelt}. We can now distinguish
between the agent's intrinsic and extrinsic perspective in this context. The world is
everything that is extrinsic from the perspective of the cognitive system,
whereas the controller, sensor and actuator signals are intrinsic to the system.

The distinction between intrinsic and extrinsic is also captured in the
representation of the sensorimotor loop as a causal or Bayesian graph (see
Fig.~\ref{fig:sml}, right-hand side). The random variables $C$, $A$, $W$, and
$S$ refer to the controller state, actuator signals, world and sensor signals,
and the directed edges reflect causal dependencies between the random variables
(see
\citep{Klyubin2004Organization-of-the-information-flow,Ay2008Information-Flows-in,Zahedi2010Higher-coordination-with}).
Everything that is extrinsic to the system is captured in the variable $W$,
whereas $S$, $C$, and $A$ are intrinsic to the system.

In this context, we distinguish between internal and external reward function (IRF, ERF) in the
following way. An ERF may access any variable, especially those that are not
available to an agent by its sensors, i.e.~anything that we summarised as the
world state $W$. An IRF may
access intrinsically available information only ($S_t$,$A_t$,$C_t$, see
Fig.~\ref{fig:sml}). We are interested in first principled model of an intrinsic
motivation, i.e.~a model that
requires as few assumptions as possible. The idea is
that IRF should not depend on a specific task but rather be a task-independent
internal driving force, which supports any task-dependent learning. This is why
we refer to it as task-independent internal motivation or drive.
This closes the discussion of embodied agents and their formalisation in terms
of the sensorimotor loop. The next section describes the information-theoretic
measures that are used in the remainder of this work.

\subsection{Predictive Information}
The predictive information (PI) \citep{Bialek1999Predictive-Information}, which
is also known as excess entropy
\citep{Crutchfield1989Inferring-statistical-complexity} and effective measure
complexity \citep{Grassberger1986Toward-a-quantitative} is defined as the mutual
information of the entire past and future of the sensor data stream:
\begin{align}
  I_{pred}(S) & := I(S_p; S_f)
\end{align}
where $S_p=\{S_1,S_2,\ldots,S_t\}$ is the entire past of the
system's sensor data at some time $t\in\mathbb{N}$ and
$S_f=\{S_{t+1},S_{t+2},\ldots\}$ its entire future. The PI
captures how much information the past carries about the future.
Unfortunately, it cannot be calculated for most applications because of technical reasons.
Hence, we use the one-step PI, which is
given by
\begin{align}
  I^*_{pred}(S)  & := I(S_{t+1}; S_t) \nonumber\\
                 & =
  \underbrace{H(S_{t+1})}_{\text{diversity}}
  \underbrace{-H(S_{t+1}|S_{t})}_{\text{compliance}},
  \label{eq:one step pi}
\end{align}
which was previously investigated in the context of EAI
\citep{Ay2008Predictive-information-and} and as a first principle learning rule
\citep{Zahedi2010Higher-coordination-with,MartiusDerAy2013}. A different
motivation for the PI is based on maximising the mutual information of
an intention state $\tilde{S}_t$, which is internally generated by the agent, and
the next sensor state
$S_{t+1}$ \citep{Ay2013An-Information-Theoretic-Approach}.
The Equation \eqref{eq:one step pi}
displays how maximising the PI affects the behaviour of a system. The
first term in Equation \eqref{eq:one step pi} leads to a maximisation of the
entropy over the sensor states. This means that the agent has to explore its
world in order to sense every state with equal probability. The second term in
Equation \eqref{eq:one step pi} states that the uncertainty of the next sensor
state must be minimal if the current sensor state is known. This means that an
agent has to choose actions which lead to predictable next sensor states. This
can be rephrased in the following way. Maximising the entropy $H(S_{t+1})$
increases the diversity of the behaviour whereas minimising the conditional
entropy $-H(S_{t+1}|S_t)$ increases the compliance of the behaviour. The result
is a system that explores its sensors space to find as many regularities in its
behaviour as possible.

For completeness we will also maximise the entropy $H(S_t)$ only and compare the
results to the maximisation of the PI. This concludes the presentation of the PI
(and entropy) as a model for a task-independent internal motivation in the
context of RL. The next section presents the utilised RL
algorithm.

\subsection{Policy Gradients with Parameter-Based Exploration (PGPE)}
We chose an episodic RL method named PGPE
\citep{Sehnke2010Parameter-exploring-policy-gradients} to investigate the effect
of the PI as an IRF, because it is not restricted to a specific class of
policies. Any policy, which can be represented by a vector $\mu\in\mathbb{R}^n$
with fixed length $n\in\mathbb{N^+}$ can be optimised by this method. In the
work presented here, we use it to learn the synaptic strengths and bias values of
neural networks with fixed structures only. Nevertheless, we can apply the
framework to other parametrisations, in particular to stochastic policies, which
is why PGPE attracted our attention for ongoing the project in which this work
is embedded.

The algorithm can be summarised in the following way (for details, see 
\citep{Sehnke2010Parameter-exploring-policy-gradients}). In each \emph{roll-out} or
episode, two policy instances are drawn from $\mu$ by adding and subtracting a
random vector $\epsilon \sim \mathcal{N}(0,\sigma)$ to it. The resulting two
policy parametrisations $\Theta^+=\mu + \epsilon$ and $\Theta^-=\mu - \epsilon$
are then evaluated and their
final rewards $r^+, r^-$ are used to determine the modifications on $\mu$ and
$\sigma$ according to the following equations\\
\begin{minipage}[c]{0.5\textwidth}
\begin{align}
  m^n &              = \max(m^{n-1}, r^{+,n}, r^{-,n})\label{eq:m}\\
  b^n &              = (1-\delta)b^{n-1} + \delta\sum_n\frac{r^{+,n}+r^{-,n}}{2}
\end{align}
\end{minipage}
\begin{minipage}[c]{0.5\textwidth}
\begin{align}
  \Delta\mu_i &    = \frac{\alpha\epsilon_i(r^+-r^-)}{2m-r^+-r^-} \label{eq:mu}\\
  \Delta\sigma_i & = \frac{\alpha}{m-b}
                     \left(\frac{r^+-r^-}{2}-b\right)
                    \left(\frac{\epsilon^2-\sigma_i^2}{\sigma_i}\right)
                    \label{eq:sigma}.
\end{align}
\end{minipage}\medskip\\
Roll-outs can be repeated several times before a learning
step is performed. Every learning step concludes a \emph{batch}. PGPE requires
an initial $\mu_\mathrm{init}$, an initial $\sigma_\mathrm{init}$, a learning
rate $\alpha$, baseline $b$,
baseline adaptation parameter $\delta$, and an
initialised maximal reward $m=m_\mathrm{init}$. We have set $\delta$ to the recommended
value of 0.1, $\mu_\mathrm{init} = 0$, and we have achieved the best results in all
experiments by setting $m_\mathrm{init}$ small enough that $m$ is definitely overwritten
in the first roll-out (see Eq.~\eqref{eq:m}). The other parameters are evaluated in each experiment,
such that the best results were achieved when no IRF was used and then fixed for
the remaining experiments.

\section{Results}
\label{sec:results}
This section presents three different experiments and their results.
The first experiment is the cart-pole swing-up, a standard control theory problem
that is also widely used in machine learning
\citep{Barto1983Neuronlike-adaptive-elements,Geva1993The-Cart-Pole,%
Doya2000Reinforcement-Learning-in-Continuous,%
Pasemann1999Evolving-structure-and-function}. The cart-pole experiment is also chosen because
balancing a pole minimises the entropy, and hence, it contradicts the
maximisation of the PI. The second experiment is the learning of a locomotion
behaviour for a hexapod and it was chosen to demonstrate the effect of the PI
maximisation on a more common, well-structured experimental setting. By
well-structured we mean that the controller, morphology, environment, and ERF
are chosen such that they result in a good hexapod locomotion without any
additional support by an IRF in only a few policy updates. The third experiment
is designed to be challenging, as it combines a high-dimensional system, an
unconventional control structure, an unsteady ERF with an unsteady environment.
We believe that these three experiments span a broad range of possible
applications for information-theoretic IRF in the context of episodic RL.

\subsection{Cart-Pole Swing-Up}
\label{sec:cart pole experiment}
The cart-pole swing-up experiment is ideal to investigate the effect of the
PI on an episodic RL task, mainly for
two reasons. First, the experiment is well-defined by a set of equations and
parameters, that are widely used in literature
\citep{Barto1983Neuronlike-adaptive-elements,%
Geva1993The-Cart-Pole,Doya2000Reinforcement-Learning-in-Continuous,%
Pasemann1999Evolving-structure-and-function}. This ensures that the results are
comparable and reproducible by others with little effort. Second,
the successful execution of the task contradicts the maximisation of the
PI. The task is to balance the pole in the centre of the
environment, and hence, to minimize the entropy of the sensor states. The
maximisation of the PI demands a maximisation of the entropy (see
Eq.~\eqref{eq:one step pi}). The remainder of this section first describes the
experimental and controller setting and then closes with a discussion of the
results.

\begin{figure}
  \begin{center}
    \includegraphics[width=0.75\textwidth]{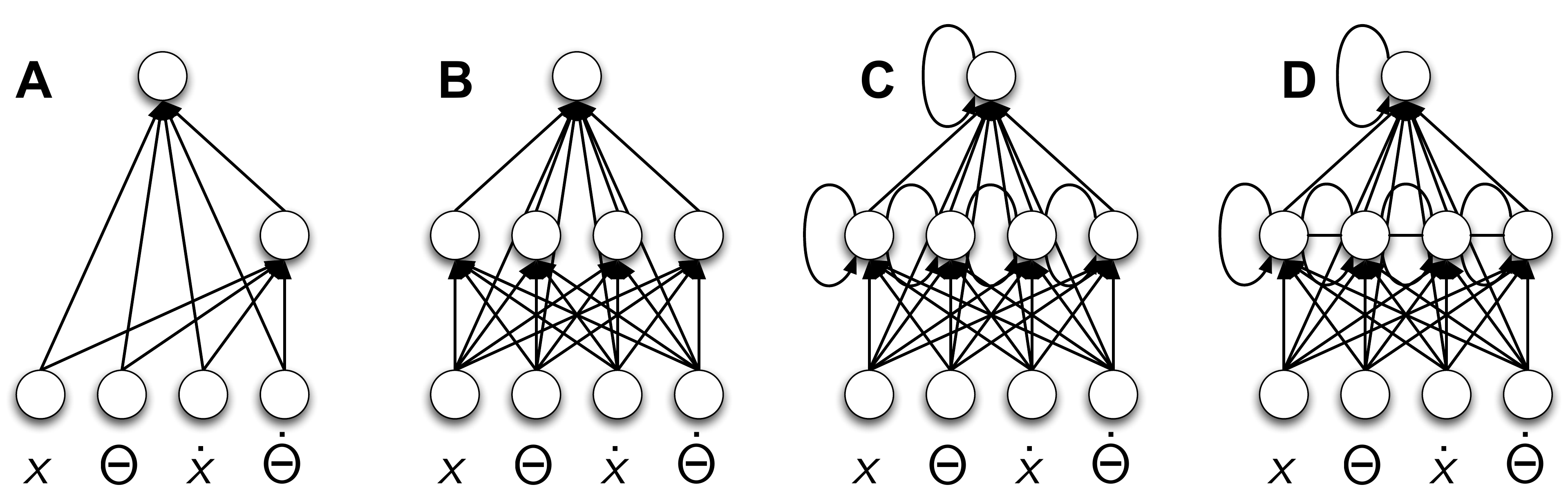}
  \end{center}
  \caption{Controller architectures for the cart-pole swing-up task.
    The input neurons are bare buffer neurons whereas the hidden and output neurons have $\tanh$ transfer-function.
    (\infig{A})~from \citep{Pasemann1999Evolving-structure-and-function};
    (\infig{B})~with 4 hidden neurons and fully connected;
    (\infig{C,D})~recurrent variations without and with lateral connections
    }\label{fig:cart-pole controller}
\end{figure}

The experiment was conducted by implementing the equations that can be found in
\citep{Barto1983Neuronlike-adaptive-elements,%
Geva1993The-Cart-Pole,Doya2000Reinforcement-Learning-in-Continuous}. The state
of the cart-pole is given by $x,\dot{x},\vartheta,\dot{\vartheta}$, which are
the position of the cart, the speed of the cart, the pole angle and the pole's
angular velocity. The cart is controlled by a force $F\in[-10N,10N]$ that is applied to
its centre of mass. The four state variables and the force define the input and
output configuration of our controllers for this task. The initial controller
(see Fig.~\ref{fig:cart-pole controller}A) was chosen from
\citep{Pasemann1999Evolving-structure-and-function}, where network structures
were evolved for the same task. To ensure that the evolved structure is not
especially unsuitable for RL, different variations were
chosen for evaluation too (see Fig.~\ref{fig:cart-pole controller}B-D). In this
approach, the input neurons are simple buffer neurons, with the identity as
transfer-function, whereas all other neurons use the hyperbolic
tangent transfer-function.

The evaluation time was set to $T=2000$ iterations, which
corresponds to 20 seconds
(c.f.~\citep{Doya2000Reinforcement-Learning-in-Continuous}). Different values,
starting from the values proposed in
\citep{Sehnke2010Parameter-exploring-policy-gradients}, for the learning rate
$\alpha\in\{\underline{0.1},0.2,0.5\}$, the initial variation
$\sigma_\mathrm{init}\in\{2,\underline{5}\}$, and the initial maximal reward
$m_\mathrm{init}\in\{\underline{-\infty}, 10, 100, 1000\}$ were evaluated in
experiments without applying an IRF to the learning of the task. The underlined
values showed the best results, and hence, are chosen for presentation here.
Each experiment consisted of $B=10000$ batches, i.e.~updates of $\mu$ and
$\sigma$ (see Eqs.~\eqref{eq:mu} and \eqref{eq:sigma}) with two roll-outs each
(i.e.~four evaluated policies $\theta^{+,-}_{1,2}$). The results are
obtained by conducting every experiment 100 times. To ensure comparability among
the experiments with different parameters and controllers, the random number
generator was initialised from a fixed set of 100 integer values for each
experiment.

The presentation of the reward function is split into two parts. The first part
handles the ERF, whereas
the second part handles the IRF. We use the terms
\emph{intrinsic/internal} and \emph{extrinsic/external} with respect to the
agent's perspective, as discussed in the previous section (see
Sec.~\ref{sec:sml}). The controller has access to the full
state of the system, and hence, the separation into internal and external is
artificial in this case. Nevertheless, we keep this terminology for
consistency, as the next experiments will reflect this distinction in a natural
way.
We denote
IRF by $R_\mathrm{in}$ and ERF by $R_\mathrm{ex}$, where a super-script is
added to distinguish between the different reward functions (PI and
entropy).

The ERF for the cart-pole swing-up task is defined such that it is not a smooth
gradient in the reward space, and therefore, does not directly guide the learning
process. The controller is only rewarded if the pole is pointing upwards and the
reward is scaled with the distance of the pole to the center of the environment,
which is given by
\begin{align}
  \Rex(t) & := \left\{
  \begin{array}{ll}
    2 - \vert x(t)\vert & \text{if }\vert\vartheta(t)\vert < 5^\circ\\
    0 & \text{otherwise.}
  \end{array}
  \right.
\end{align}

The IRF is calculated at the end of each episode based on the recordings of the
pole angles $\{S_t=\vartheta(t)|t=1,2,\ldots,T\}$. We use a discrete-valued
computation of the PI, and hence, the data is binned prior to the calculation.
All IRFs are normalised with respect to their theoretical upper bound of
$I(S_{t+1};S_t)\leq H(S_t)\leq\log|S|$ (see
\citep{Cover2006Elements-of-Information}). This leads to the two following IRFs: \\
\begin{align}
  \Rpi & := |I(S_{t+1};S_t)|\qquad\textrm{and} \qquad \Rh   := |H(S_t)|.
\end{align}
The overall reward functions are then given by
\begin{align}
  R^\mathrm{PI} & := \sum_{t=1}^T \Rex(t) + \beta(\gamma) \Rpi, &
  R^\mathrm{H} & := \sum_{t=1}^T \Rex(t) + \beta(\gamma) \Rh, &
  \beta(\gamma) = \gamma \cdot T \cdot\max_{x,\vartheta,t}\left\{\Rex(t)\right\}
  \label{eq:beta}
\end{align}
where $\beta(\gamma)$ is a factor to scale the IRF with respect to the
maximal possible value of the ERF\@. This allows us to compare the effects of
$R_\mathrm{in}^{\mathrm{PI}}$  and
$R_\mathrm{in}^{\mathrm{H}}$ across different experiments.

\begin{figure}
  \begin{center}
    \includegraphics[width=\textwidth]{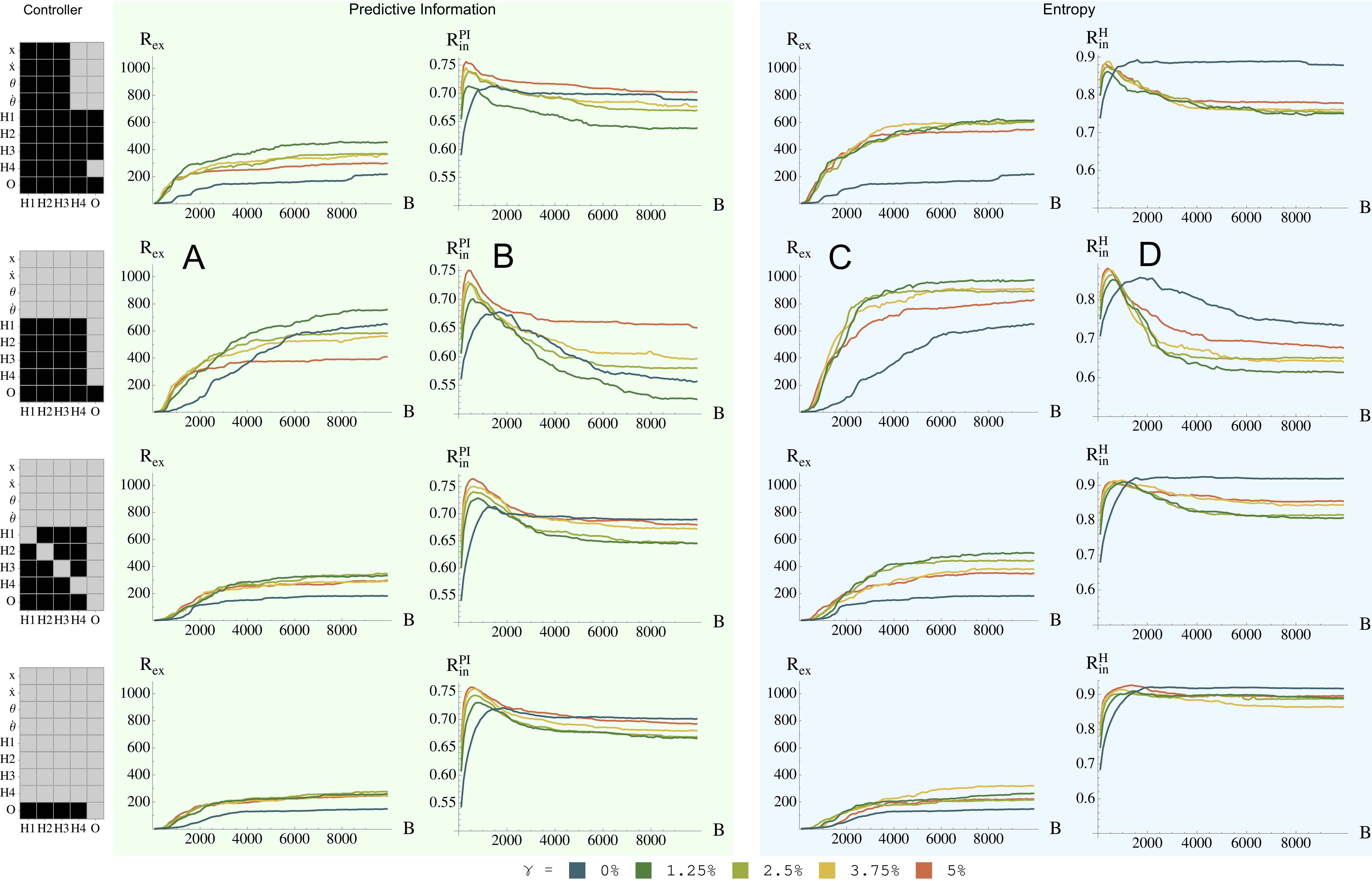}
  \end{center}
  \caption{Results for cart-pole experiments.
    Each row shows the results for one controller architecture, see
    Fig.~\ref{fig:cart-pole controller}. The corresponding connection matrix is
    provided in the first column (gray: connection, black: no
    connection). For
    simplicity only the row for the second controller is discussed in detail.
    \infig{(A,B)}~ERF and IRF for PI maximisation -- small values of $\gamma>0$
    are advantageous.  \infig{(C,D)}~ERF and IRF for entropy maximisation -- all
  values of $\gamma>0$ have positive effect.}
  \label{fig:cart-pole results}
\end{figure}

The results are discussed only for the fully connect feed-forward network (see
Fig.~\ref{fig:cart-pole results}A--D) in detail as this controller shows the
most distinguishable results with respect to the variation of the IRF scaling
parameter $\gamma\in\{0\%,1.25\%,2.5\%,3.75\%,5\%\}$. It is important to note
that the plots only show the averages of the 100 experiments and not the standard
deviation for the following reason. Few controllers succeed early, others later
during the process. Due to the unsteady ERF the resulting standard
deviation is very large, as those controllers that succeed receive significantly higher reward
compared to those not succeeding (which remain close to zero, as a rotational
behaviour is not permitted). We intentionally chose an unsteady ERF,
that returns zero for almost all states, and hence, we know beforehand, that the
standard deviation is large and no further information is provided if it is
plotted.

Figures~\ref{fig:cart-pole results}A and \ref{fig:cart-pole results}B show the
progress of the ERF $R^\mathrm{PI}_\mathrm{ex}$ and IRF
$R_\mathrm{in}^\mathrm{PI}$ for the PI maximisation. It is shown that there is a
significant speed-up in learning during the first 4000 batches for all
$\gamma>0\%$ (see Fig.~\ref{fig:cart-pole results}A). At this point in time the
average ERF of $\gamma=0\%$ succeeds that of $\gamma=5\%$. After approximately
5000 batches the ERF for $\gamma=2.5\%$ and $\gamma=3.75\%$ are very close to or
slightly succeeded by the ERF for $\gamma=0\%$, whereas the ERF for
$\gamma=1.25\%$ remains higher. The conclusion from this experiment is that
small values of $\gamma<5\%$ are beneficial in this learning task as less
batches are required to solve this task and the
asymptotic learning performances are almost identical to $\gamma=0\%$.
The results, however, are not significant and the choice of $\gamma$ is critical.
This leads to the conclusion that the one-step PI is not significantly beneficial
in the learning of this task.

Figures~\ref{fig:cart-pole results}C and \ref{fig:cart-pole results}D show the
progress of the ERF $R^\mathrm{H}_\mathrm{ex}$ and IRF $R_\mathrm{in}^\mathrm{H}$ for the
entropy maximisation. The results show a different picture. Any parameter
$\gamma>0\%$ speeds up the learning and improves the overall performance. The
comparison of entropy and PI is addressed in the discussion again.


\subsection{Hexapod Locomotion}
\label{sec:hexapod:locomotion}
If a specific task should be learned by an embodied agent, it is more common to
choose an environment, morphology, control structure and a smooth ERF which are
well-suited for the desired task. In order to investigate which effect the PI
has on such a well-defined learning task, the set-up of the experiment presented
in this section is chosen such that all components are known to work well if
there is no IRF present. The goal is to learn a locomotion behaviour of a
hexapod, where the maximal deviation angles ensure that it
cannot flip over. The controller is known to perform well in a similar
task \citep{Markelic2007An-Evolved-Neural} and its modularity significantly
reduces the number of parameters that must be learned. The ERF defines a smooth
gradient in the reward space, ensuring that small changes in the controller
parameters show an immediate effect in the ERF. The environment is an 
even plane without any obstacles. 

\begin{figure}[t]
  \begin{center}
    \includegraphics[width=\textwidth]{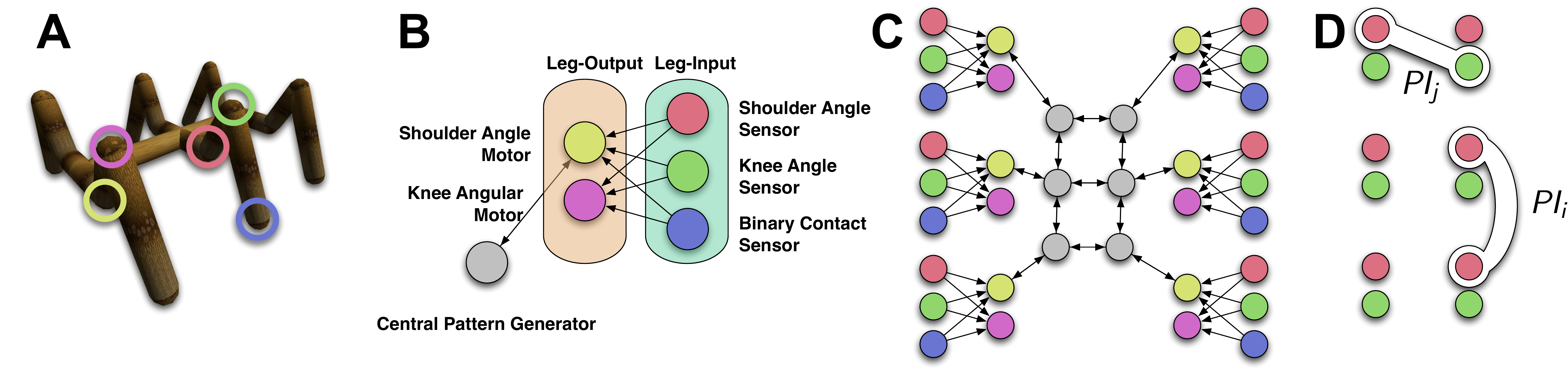}
  \end{center}
  \caption{Hexapod for locomotion task and controller set-up.
    (\infig{A})~Hexapod robot with marked actuated joints and sensors;
    (\infig{B})~leg module of controller;
    (\infig{C})~entire controller;
    and (\infig{D})~schematic pairings for PI and entropy calculation.}
  \label{fig:hexapod set-up}
\end{figure}

The experimental platform (see Fig.~\ref{fig:hexapod set-up}) is a
hexapod, with 12 degrees of
freedom (two actuators in each leg) and with 18 sensors (angular positions of the
actuators and binary foot contact sensors). The two actuators of each leg are
positioned in the shoulder (Thorax-Coxa or ThC joint) and in the knee
(Femur-Tibia or FTi joint) of the walking machine, similar to the morphology
presented in \citep{Twickel2011Deriving-neural-network}. We omit the second
shoulder-joint (CTr) because it is not required for locomotion.
Each joint accepts the desired angular position as its input and returns the
actual current angular position as its output.
The simulator YARS \citep{Zahedi2008YARS:-A-Physical} was used for all
experiments conducted in this section.

Different values for the PGPE parameters were evaluated. The best results for
$\gamma=0$ (see Eq.~\eqref{eq:beta}) were achieved with $\sigma_\mathrm{init}=2$ and
$\alpha=0.1$. To ensure comparability with the previous experiment, two roll-outs
were chosen here, although it is not required to obtain the following results.
The evaluation time was set to $T=1000$ and $B=250$ batches were sufficient to observe a
convergence of the policy parameters $\mu$. The values for $\gamma$ were chosen
from the previous experiment.

The ERF is calculated once at the end of each episode and it is defined as the
euclidean distance between the hexapod at time $T$ and its initial position $(0,0)$
projected onto the $xy$-plane:
\begin{align}
  \Rex & := \sqrt{x^2_T + y^2_T},
\end{align}
where $(x_T,y_T)$ are the coordinates of the centre of the robot
in world coordinates at time $t=T$.

The IRF is calculated differently compared to the
previous experiment. In a high-dimensional system as the hexapod, it is not
possible to compute the PI of the entire system with a reasonable
effort, as the computational cost of $I(S_t;S_{t+1})$ grows exponentially for
every new sensor. It would be natural to reduce the computational cost by
calculating the PI based on a model of the morphology, but this would violate
our claim that the PI incorporates the morphology without
the need of explicitly modelling it. Hence, we decided to
use the following method to approximate the PI and the entropy $H$ (see
Fig.~\ref{fig:hexapod set-up}D). Let $S_i(t), i=1,2,\ldots,12$, be the angular position
sensors for the 12 actuators. We then chose two sensors $k,l$ with $1\leq k,l
\leq 12, k\not=l$, randomly from the
12 possibles sensors, and calculated
\begin{align}
  PI_u & := I(S_k(t+1),S_l(t+1);S_k(t),S_l(t)) &
  H_u & := H(S_k(t),S_l(t)).\label{eq:locomotion:piu}
\end{align}
The overall PI and entropy are then calculated as the sum of $n$ randomly chosen
$PI_u$ and $H_u$ pairings, with the additional constraint that each sensor pair
$k,l$ appears only once in the approximations. The resulting IRFs are then given
by:
\begin{align}
  \Rpi & := \sum_{u=1}^n PI_u \qquad \textrm{and}\qquad \Rh  := \sum_{u=1}^n H_u,\label{eq:locomotion:pi}
\end{align}
where $n$ is the number of pairings. For $n>20$ no
difference was found for the approximated PI, which is why $n=20$ was chosen for
the remainder of this work.

The overall reward functions are then given by:
\begin{align}
  R^\mathrm{PI} & := \Rex + \beta(\gamma) \Rpi &
  R^\mathrm{H}  & := \Rex + \beta(\gamma) \Rh
\end{align}
where $\beta(\gamma)$ is defined as in the cart-pole swing-up experiment (see
Eq.~\eqref{eq:beta}).

A common recurrent neural network central pattern generator layout is chosen,
which can also be found in
literature
\citep[e.g.][]{Campos2010Hexapod-locomotion:-A-nonlinear,Twickel2011Deriving-neural-network,Markelic2007An-Evolved-Neural},
thereby using the same neuron model as in the cart-pole experiment (see above).
As all legs in the hexapod are morphologically equivalent, only
the synaptic weights of one leg controller are open to parameter adaptation in
the PGPE algorithm. The values are then copied to the other leg controllers.
This reduces the number of parameters for the entire controller to 32 (see
Figs.~\ref{fig:hexapod set-up}B and \ref{fig:hexapod set-up}C).

\begin{figure}[t]
  \begin{center}
    \includegraphics[width=\textwidth]{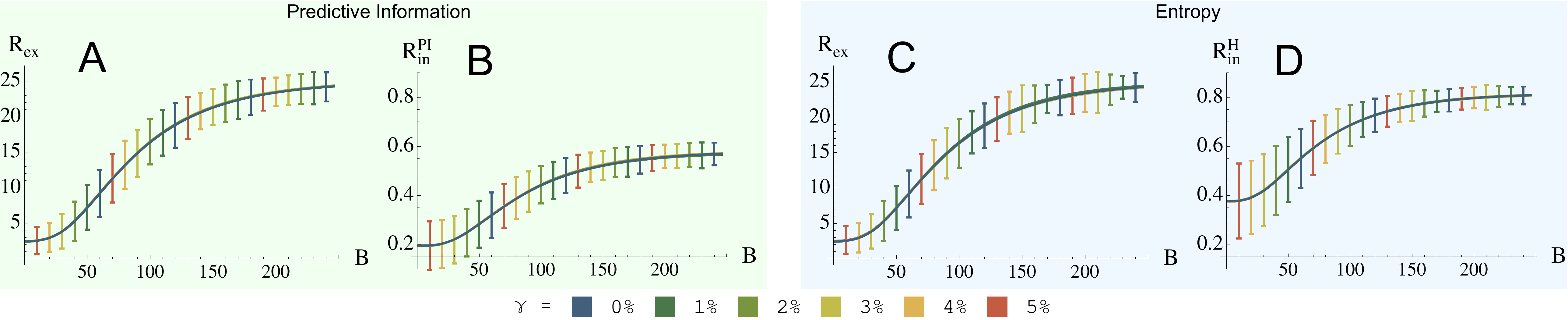}
  \end{center}
  \caption{Results for hexapod locomotion task.
    ERF and IRF with PI maximisation (\infig{A,B}) and entropy maximisation (\infig{C,D}).
    No significant effect is observed.
  }
  \label{fig:hexapod locomotion}
\end{figure}

The results (see Fig.~\ref{fig:hexapod locomotion}) show that neither the
PI nor the entropy have a noticeable effect on the learning performance. The mean
values of the 100 experiments for each parameter as well as the standard
deviation are almost identical. This point will be addressed in the discussion of this
work (see Sec.~\ref{sec:discussion}).

\subsection{Hexapod Self-Rescue}
The third experiment is designed to combine and extend the two previous
experiments. It combines them as a high-dimensional morphology, similar to that
used in the locomotion experiment, is trained with an unsteady ERF, which is
similar to that used in the cart-pole experiment. It extends the previous
experiments as the number of parameters in the controller is a magnitude larger
and because an unconventional control structure is used for the desired task.
The most distinctive difference to the previous experiments is the non-trivial
environment. The next paragraphs will explain the experimental set-up in detail
before the section closes with a discussion of the results.

\begin{figure}
  \centering
  \begin{tabular}{cc}
    \infig{A} & \infig{B} \\
  \includegraphics[width=0.3\linewidth]{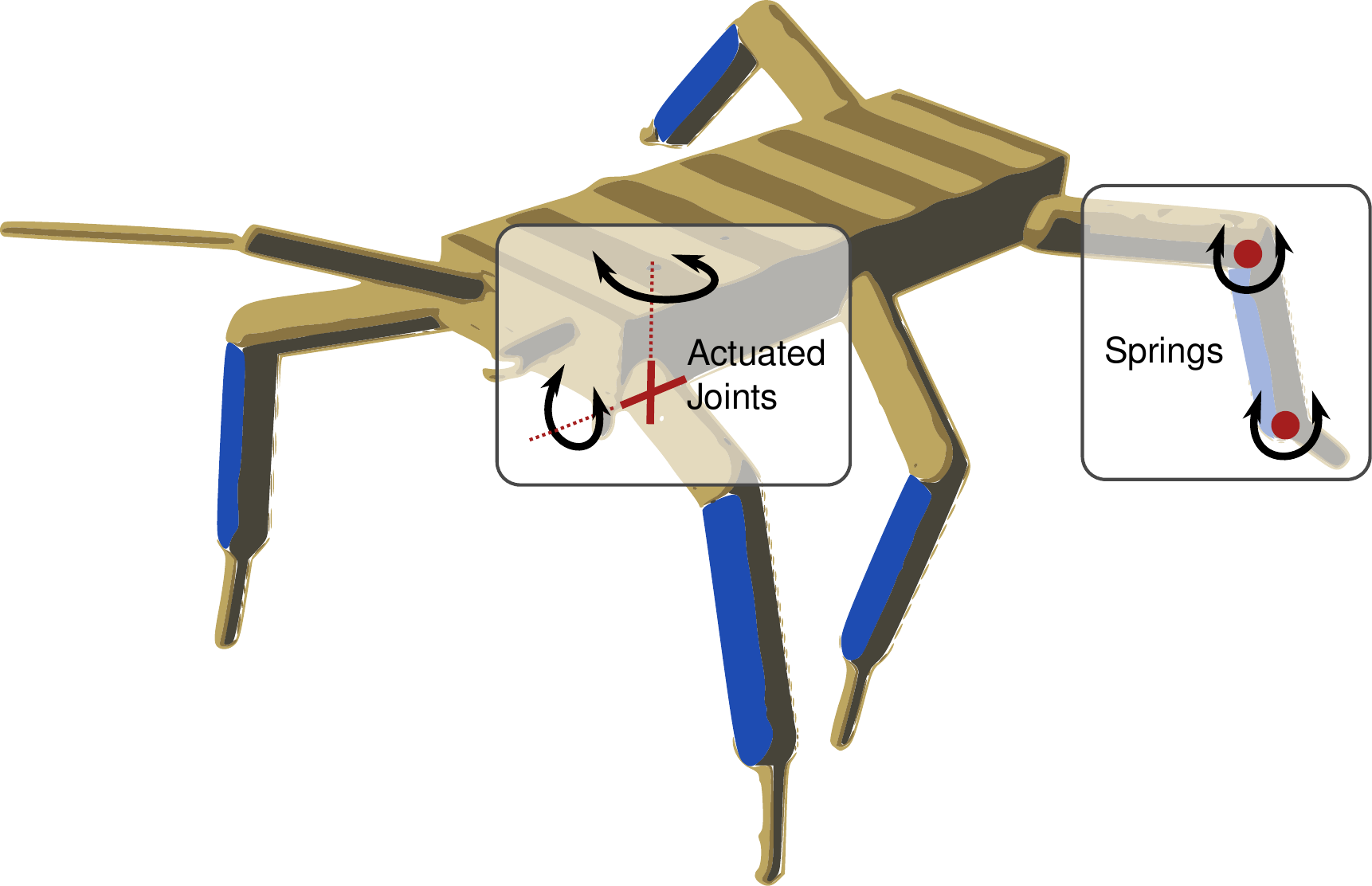}&
  \includegraphics[width=0.55\linewidth]{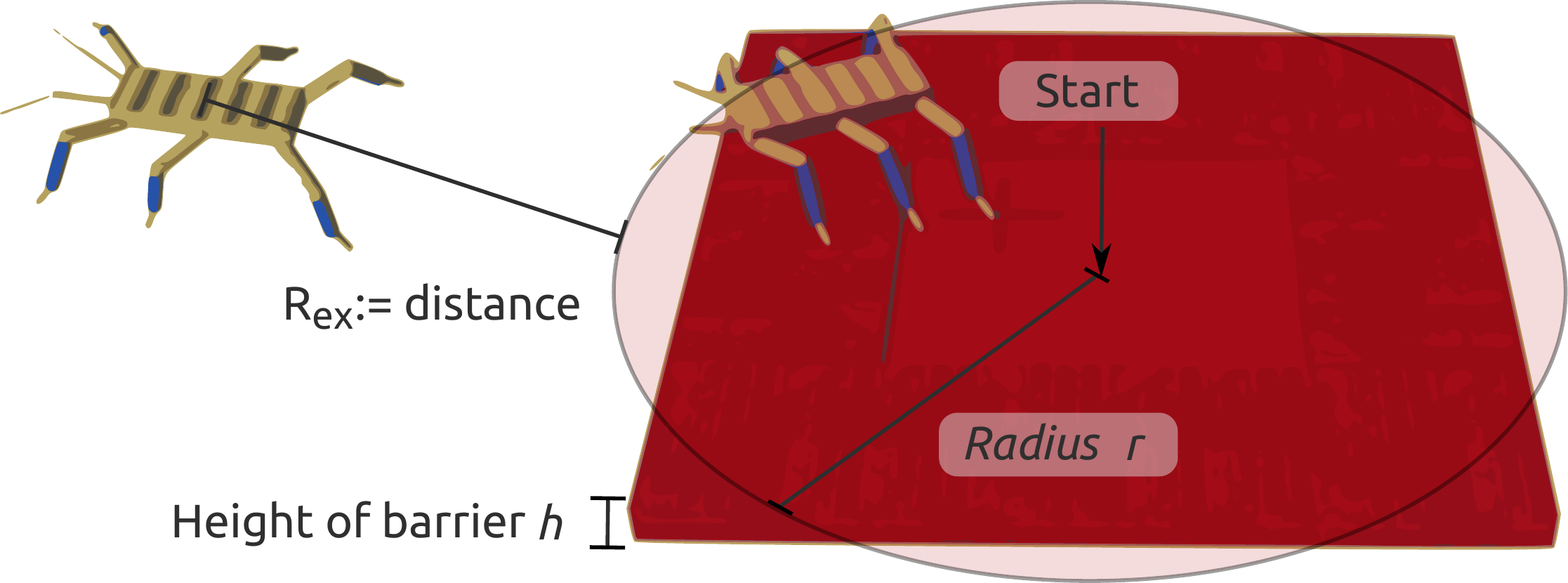}
  \end{tabular}
  \caption{Hexapod robot for self-rescue and the experimental setup.
    (\infig{A})~The robot has 6 legs where the hind legs are 10\% larger
    than the other legs. Each leg has two active
    DoF in the
    hip joint and one passive DoF in both the knee and the ankle
    joint equipped with a spring.
    Additionally the whiskers have each two spring-joints.
    (\infig{B})~The robot starts in the centre of the trap with a
    certain barrier height and has to escape from it. The reward is
    the distance from the outside of the trap or zero otherwise.
  }
\label{fig:hexapod:rescue:setup}
\end{figure}

We used the simulated hexapod robot of the
\textsc{LpzRobots} simulator~\citep{lpzrobots12}. The hexapod has 12 active and
16 passive degrees of freedom (see Fig.~\ref{fig:hexapod:rescue:setup}). The
active joints take the desired next angular position as their input and deliver
the current actual angular position as their output. The controller is a fully
connected one-layer feed-forward neural network without lateral connections and
the hyperbolic transfer function $a_{t+1} = \tanh(Ws_t+v)$, where $a_{t+1}$ and
$s_t$ are the next action and the current sensor values, $W$ is the connection
matrix, and $v$ is the vector of biases. The resulting controller is parameterised by
156 parameters, 144 for the synaptic weights and 12 for the bias values.
Note, that the controller is generic and has no a priori structuring or other
robot-specific details.

\begin{figure}
  \centering
  \includegraphics[width=\textwidth]{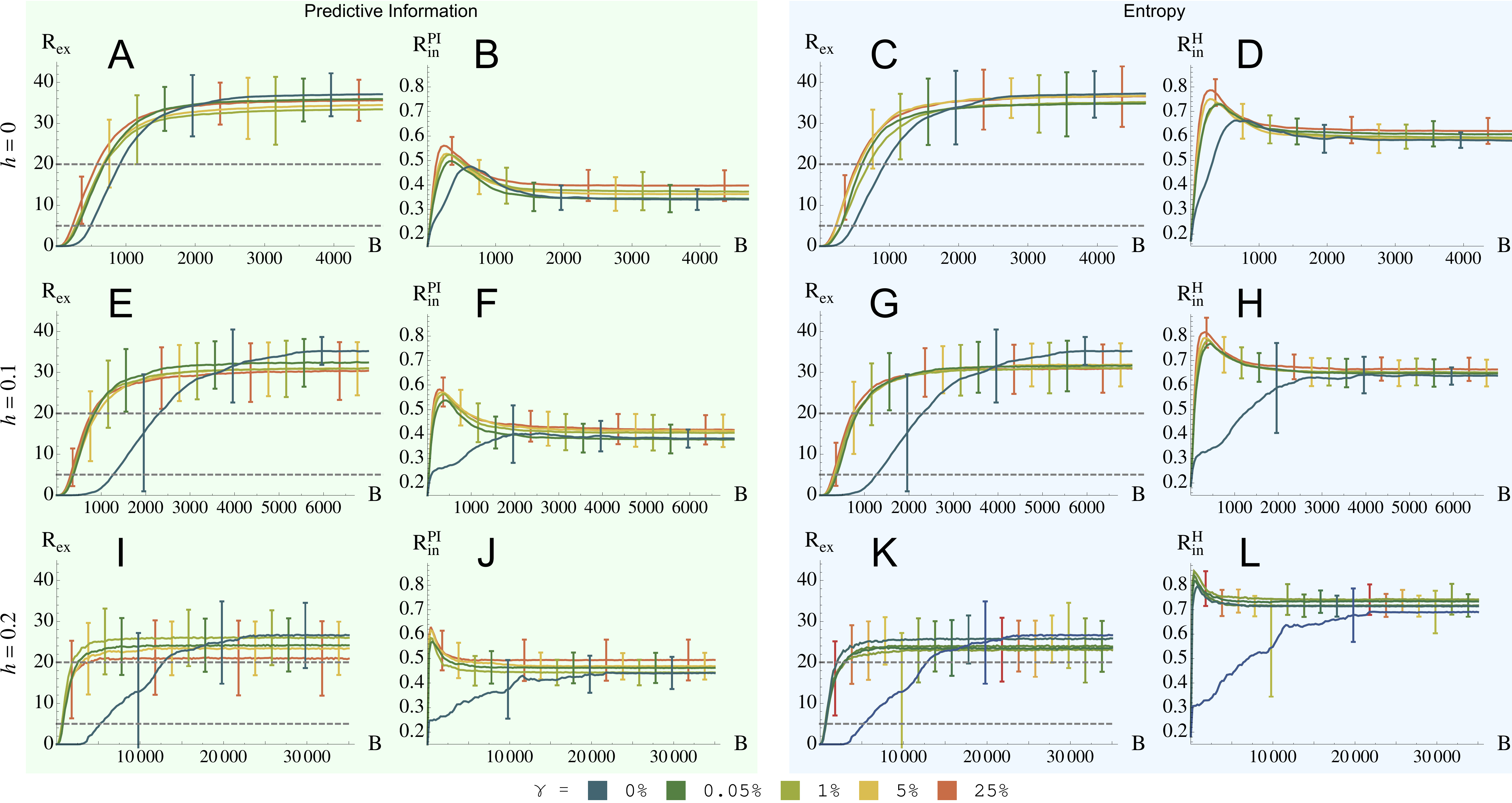}
  \caption{Performance in the self-rescue task depending
    on the internal reward type and factor $\gamma$.
    Plotted are the ERF and the IRF in case of PI
    (\infig{A},\infig{B},\infig{E},\infig{F},\infig{I},\infig{J})
    and entropy (\infig{C},\infig{D},\infig{G},\infig{H},\infig{K},\infig{L})
    over the number of
    batches for different values of $\gamma$ and barrier heights $h$:
    (\infig{A}--\infig{D}) no barrier ($h=0$),
    (\infig{E}--\infig{H}) low barrier ($h=0.1$) and
    (\infig{I}--\infig{L}) high barrier ($h=0.2$).
    For each value of $\gamma$ the mean and standard deviation
     of 30 experiments are displayed.
    In all cases a speed-up in learning is achieved with IRF, however, the
    asymptotic performance is worse.
  }
\label{fig:rescue:results}
\end{figure}

The task for the hexapod is to rescue itself from a trap. For this purpose, it
is placed in a closed rectangular arena (see
Fig.~\ref{fig:rescue:results}). The difficulty of the task is
determined by the height of the arena's walls, denoted by $h\in\{0.0\textrm{m},
0.1\textrm{m}, 0.2\textrm{m}\}$  (see
Fig.~\ref{fig:hexapod:rescue:setup}). For comparison, the length of the lower leg
(up to the knees) is $0.45$m. The size-proportion of the robot and the trap
can be seen in Fig.~\ref{fig:hexapod:rescue:setup}B.

The ERF is given by
\begin{align}
  \Rex & :=
   \begin{cases}
     \sqrt{x^2_T + y^2_T} - r &\text{if } \sqrt{x^2_T + y^2_T} - r > 0\\
     0 &\text{otherwise,}
   \end{cases}
\end{align}
where $r$ is the radius of the trap (Fig.~\ref{fig:hexapod:rescue:setup}) and
$(x_T,y_T)$ is the position of the centre of the robot in world coordinates at
the end of a roll-out ($t=T$). The IRFs and overall reward functions are
identical to those used in the previous experiment (see
Eqs.~\eqref{eq:locomotion:piu} and \eqref{eq:locomotion:pi}).

As before, the performance of PGPE with $\gamma=0$ for different values for
$\sigma_{\textrm{init}}$ and $\alpha$ were evaluated, and the best are chosen
for presentation here, which are $\sigma_{\textrm{init}}=2$ and $\alpha=0.5$. A
different learning rate $\alpha_\sigma = 0.05$ was chosen for the update of
$\sigma$ (see Eq.~\eqref{eq:sigma}). Each episode consisted of $T=1250$
iterations ($25$s) with one roll-out per episode. A total of $B=5000, 7000$, and
$35000$ batches were conducted for the different heights $h$.

We compare the performance for different values of the IRF factor
$\gamma\in\{0\%, 0.05\%, 1\%, 5\%, 25\%\}$ and performed 30 experiments for each
setting. Figure~\ref{fig:rescue:results} displays the results. As for the
cart-pole experiment, the plots for the PI and entropy in
Fig.~\ref{fig:rescue:results} report a clear picture of an exploration phase
(high value) followed by an exploitation phase (lower value).

To compare the results, we set two threshold values at $\Rex=5$ and $\Rex=20$
which refer to a $5$m and $20$m distance between the hexapod and the walls of
the arena. The first threshold reflects a successful learning of the task,
because it means that hexapod reliably escapes the arena. The second threshold
represents the case when in addition also a high locomotion speed is achieved
after a successful escape. For the simplicity of argumentation, we compare two
cases, i.e.~$\gamma=0\%$ and $\gamma=1\%$. If there is no wall ($h=0$m) the
system with IRF $\gamma=1\%$ requires only half the amount of batches compared
to no IRF (250 batches vs.~500 batches, see Figs.~\ref{fig:rescue:results}A and
\ref{fig:rescue:results}C). For the arena with a medium height ($h=0.1$m), the
learning success speed ratio increases to approximately three (350 batches
vs.~1100 batches, see Figs.~\ref{fig:rescue:results}E and
\ref{fig:rescue:results}F). The results are decisive for the arena with high
walls ($h=0.2$m), as the system with IRF requires about 1000 batches on average
compared to the 5000 batches on average that a required by the systems without
IRF (see Figs.~\ref{fig:rescue:results}I and \ref{fig:rescue:results}K).

This leads to the conclusion that both, PI and
entropy, are beneficial if the short-term learning success is of the primary
interest.
However, the asymptotic learning success of those hexapods with IRF is either equal or
lower compared to those without an IRF in all experiments. This is valid for the
one-step PI and for the entropy.
Thus, both are not necessarily beneficial if the long-term, asymptotic
learning performance in an episodic policy gradient setting is important.

\section{Discussion} \label{sec:discussion}
This paper discussed the one-step PI \citep{Bialek1999Predictive-Information} as an
information-driven intrinsic reward in the context of an episodic policy
gradient method. The reward is considered to be intrinsic,
because it is
task-independent and it relies only on the information of the sensors of an
agent, which, by definition, represent the agent's intrinsic view on the world.
We chose the maximisation of the one-step PI
 as an IRF, because it
has proved to encourage behaviours which show properties
of morphological computation without the need to model the morphology
\citep{Zahedi2010Higher-coordination-with}.

The IRF was linearly combined with a task-dependent ERF in an
episodic RL setting. Specifically, PGPE
\citep{Sehnke2010Parameter-exploring-policy-gradients} was chosen as
RL method, because it allows to learn arbitrary policy
parametrisations. Within this set-up, three different types of experiments were
performed. The following paragraph will summarise the results before
they are discussed.

The first experiment was the learning of the cart-pole swing-up task. Four
controllers were evaluated of which three were less successful and one showed
good results. The ERF was designed to be difficult to maximise without the IRF,
and the task contradicted the maximisation of the entropy and PI. The best
controller did not show a significant improvement of the learning performance
with respect to its asymptotic behaviour. An improvement could only be observed
during the first learning steps. Moreover, the choice of the linear
combination factor $\gamma$ is critical. For all controllers a minor and not
significant improvement is observable. In case of the entropy maximisation, any
factor $\gamma>0\%$ showed an improvement in learning speed and learning
performance.

A locomotion behaviour was learned for a hexapod in the
second experiment. The entire set-up used well-known components for the
environment, modular controller, ERF, and morphology so
that the task was solved without IRF in only a few
learning steps. No effect of the PI and entropy was
observed.

The third experiment combined the previous two and extended them by a
non-trivial environment. A hexapod had to escape from a trap and was
only rewarded outside of it. The results showed no significant
difference between the PI and the entropy as IRFs.
The learning speed was significantly improved by both IRFs with increasing
difficulty of the task. The asymptotic performance was either equal or worse
when an IRF was introduced.

The hexapod locomotion experiment teaches us, that the information-theoretic
reward functions (PI and entropy) has no effect in well-defined experimental
set-ups.

The cart-pole and the hexapod self-rescue experiments teach us that the maximal
values of the IRF should be around one percent of the maximal ERF value to
improve the
learning speed and learning performance in the short-term. The asymptotic
behaviour is either not or negatively effected by the one-step PI. The cart-pole
experiment indicates that maximising the entropy is superior to maximising the
PI, whereas the hexapod self-rescue does not show such a clear picture. The
success of the entropy in both experiments is explained by the ERFs. Due to
their nature, random changes in the policy parameters are unlikely to result in
changes in the ERF during the first batches. Hence, maximising the entropy
results in an exploration until the ERF is triggered.

The PI, defined as the entropy over the sensor states subtracted by the
conditional entropy of consecutive sensor states does not result in superior
results for the cart-pole compared to just using the entropy for the following
reason. In this set-up, the morphology and environment are very simple and
deterministic, and therefore, do not produce any noise or other uncertainties in
the sensor data stream. The uncertainty about the next possible angular position
of the pole is small, if the current pole position is known. In other words, the
cart-pole system is regular by definition and no further regularities can be
found by maximising the PI. We speculate that the conditional entropy, which
cannot be reduced by the learning in this setting, 
dampens the exploration effect of the entropy term in the PI maximisation. For
the hexapod rescue experiment, the situation is different. There is an
uncertainty about the next sensor state, given the current sensor state which
result from the morphology and the construction of the arena. The PI
maximisation is able to find regularities which can then be exploited to
maximise the ERF in the RL setting.

The results contradict our intuition, as the one-step predictive
information has shown good results when applied as an information-driven
self-organisation principle in the context of embodied artificial intelligence
\citep{Zahedi2010Higher-coordination-with,MartiusDerAy2013}. The intuitively
plausible next step was to guide the information-driven self-organization towards solving a
goal by combining it with an external reward signal in an reinforcement
learning context. The approach evaluated in this paper was to linearly combine
the PI with and external reward signal in an episodic policy gradient learning.
If anything, then the PI showed positive short-term results, if
the world was considerably probabilistic and if the external reward was sparse.
Compared to no intrinsic reward the PI showed negative results for its
asymptotic behaviour. The performance of the PI was either equal or worse
compared to the entropy in all cases.
This leads to the conclusion that research in the context of information-driven
intrinsic rewards and reinforcement learning should be carried out in other
directions, which are briefly described in the final paragraph.

We have used a constant combination factor $\gamma$ for all experiments
presented in this work. It is known from general learning theory that a decaying
learning rate is required for the convergence of a system. We chose not to use
a decaying learning factor, because this means that the internal drive is slowly
dampened until its effect is neglectable (at least in a technical application).
This would contradict the idea of motivation-driven and open-ended learning of
embodied agents.
However, the results of our present paper reveal a disadvantage of this
approach in the asymptotic limit, and therefore, suggest, contrary to our
original thoughts, to pursue a strategy with a
decaying combination factor.
The second possible
modification of this approach is to exchange the linear combination of the
internal and external reward by a non-linear function, of which 
multiplicative and exponential functions are two examples. Third, using a
gradient of the PI instead of a random exploration in the context of RL is a
promising approach that is currently investigated. In this approach, we will use
a gradient on an estimate of the PI and not the error of a predictor as in
e.g.~\citep{Schmidhuber1991Curious-Model-Building-Control}.
Fourth, we will continue to
evaluate other information-theoretic measures in the context of task-dependent
learning with the support of information-driven intrinsic motivation.
In addition to using correlation measures, such as the mutual information, we
believe that causal measures in the sensorimotor loop
\citep{Ay2013An-Information-Theoretic-Approach}, such as the measure considered
in \citep{Zahedi2013Quantifying-Morphological-Computation}, are good candidates
for future research in this field.

\section*{Acknowledgements}
This work was funded by the German Priority Program \emph{Autonomous Learning}
(DFG-SPP 1527). We would like to thank Chrisantha Fernando and the anonymous
reviewers for their very helpful comments.

\bibliographystyle{plainnat}

\begin{thebibliography}{42}
\expandafter\ifx\csname natexlab\endcsname\relax\def\natexlab#1{#1}\fi
\expandafter\ifx\csname url\endcsname\relax
  \def\url#1{{\tt #1}}\fi

\bibitem[Ay et~al.(2008)Ay, Bertschinger, Der, G{\"u}ttler, and
  Olbrich]{Ay2008Predictive-information-and}
N.~Ay, N.~Bertschinger, R.~Der, F.~G{\"u}ttler, and E.~Olbrich.
\newblock Predictive information and explorative behavior of autonomous robots.
\newblock {\em European Physical Journal B}, 63\penalty0 (3):\penalty0
  329--339, 2008.

\bibitem[Ay and Polani(2008)]{Ay2008Information-Flows-in}
N.~Ay and D.~Polani.
\newblock Information flows in causal networks.
\newblock {\em Advances in Complex Systems}, 11\penalty0 (1):\penalty0 17--41,
  2008.

\bibitem[Ay and Zahedi(2013)]{Ay2013An-Information-Theoretic-Approach}
N.~Ay and K.~Zahedi.
\newblock An information theoretic approach to intention and deliberative
  decision-making of embodied systems.
\newblock In {\em Advances in cognitive neurodynamics III}. Springer,
  Heidelberg, 2013.

\bibitem[Barto et~al.(2004)Barto, Singh, and
  Chentanez]{Barto2004Intrinsically-motivated-learning}
A.~G. Barto, S.~Singh, and N.~Chentanez.
\newblock Intrinsically motivated learning of hierarchical collections of
  skills.
\newblock In {\em Proc.~of 3rd Int.~Conf.~Development Learn.}, pages 112--119,
  2004.

\bibitem[Barto et~al.(1983)Barto, Sutton, and
  Anderson]{Barto1983Neuronlike-adaptive-elements}
A.~G. Barto, R.~S. Sutton, and C.~W. Anderson.
\newblock Neuronlike adaptive elements that can solve difficult learning
  control problems.
\newblock {\em IEEE Transactions on Systems, Man, and Cybernetics},
  SMC-13:\penalty0 834--846, 1983.

\bibitem[Bellman(2003)]{Bellman2003Dynamic-Programming}
R.~E. Bellman.
\newblock {\em Dynamic Programming}.
\newblock Dover Publications, Incorporated, 2003.

\bibitem[Bialek et~al.(2001)Bialek, Nemenman, and
  Tishby]{Bialek1999Predictive-Information}
W.~Bialek, I.~Nemenman, and N.~Tishby.
\newblock Predictability, complexity, and learning.
\newblock {\em Neural Computation}, 13\penalty0 (11):\penalty0 2409--2463,
  2001.

\bibitem[Brooks(1991)]{Brooks1991Intelligence-Without-Reason}
R.~A. Brooks.
\newblock Intelligence without reason.
\newblock In John Myopoulos and Ray Reiter, editors, {\em Proceedings of the
  12th International Joint Conference on Artificial Intelligence ({IJCAI}-91)},
  pages 569--595, Sydney, Australia, 1991. Morgan Kaufmann publishers Inc.: San
  Mateo, CA, USA.

\bibitem[Campos et~al.(2010)Campos, Matos, and
  Santos]{Campos2010Hexapod-locomotion:-A-nonlinear}
R.~Campos, V.~Matos, and C.~Santos.
\newblock Hexapod locomotion: A nonlinear dynamical systems approach.
\newblock {\em Conference Of Ieee Industrial Electronics. Proceedings}, pages
  1546--1551, 11 2010.

\bibitem[Clark(1996)]{Clark1996Being-There:-Putting}
A.~Clark.
\newblock {\em Being There: Putting Brain, Body, and World Together Again}.
\newblock MIT Press, Cambridge, MA, USA, 1996.

\bibitem[Cover and Thomas(2006)]{Cover2006Elements-of-Information}
T.~M. Cover and J.~A. Thomas.
\newblock {\em Elements of Information Theory}, volume 2nd.
\newblock Wiley, Hoboken, New Jersey, USA, 2006.

\bibitem[Crutchfield and
  Young(1989)]{Crutchfield1989Inferring-statistical-complexity}
J.~P. Crutchfield and K.~Young.
\newblock Inferring statistical complexity.
\newblock {\em Phys. Rev. Lett.}, 63\penalty0 (2):\penalty0 105--108, Jul 1989.

\bibitem[Cuccu et~al.(2011)Cuccu, Luciw, Schmidhuber, and Gomez]{cuccu2011}
G.~Cuccu, M.~Luciw, J.~Schmidhuber, and F.~Gomez.
\newblock Intrinsically motivated evolutionary search for vision-based
  reinforcement learning.
\newblock In {\em Proceedings of the 2011 IEEE Conference on Development and
  Learning and Epigenetic Robotics IEEE-ICDL-EPIROB}. IEEE, 2011.

\bibitem[Dayan and
  Balleine(2002)]{Dayan2002Reward-Motivation-and-Reinforcement}
P.~Dayan and B.~W. Balleine.
\newblock Reward, motivation, and reinforcement learning.
\newblock {\em Neuron}, 36:\penalty0 285--298, 2002.

\bibitem[Der and Martius(2012)]{Der2012The-Playful-Machine:-Theoretical}
R.~Der and G.~Martius.
\newblock {\em The Playful Machine: Theoretical Foundation and Practical
  Realization of Self-Organizing Robots}.
\newblock Cognitive Systems Monographs. Springer, 2012.

\bibitem[Doya(2000)]{Doya2000Reinforcement-Learning-in-Continuous}
K.~Doya.
\newblock Reinforcement learning in continuous time and space.
\newblock {\em Neural Computation}, 12\penalty0 (1):\penalty0 219--245, 2000.

\bibitem[Geva and Sitte(1993)]{Geva1993The-Cart-Pole}
S.~Geva and J.~Sitte.
\newblock The cart pole experiment as a benchmark for trainable controllers.
\newblock {\em IEEE Control Systems Magazine}, 13\penalty0 (5):\penalty0
  40--51, 1993.

\bibitem[Grassberger(1986)]{Grassberger1986Toward-a-quantitative}
P.~Grassberger.
\newblock Toward a quantitative theory of self-generated complexity.
\newblock {\em International Journal of Theoretical Physics}, 25\penalty0
  (9):\penalty0 907--938, 09 1986.

\bibitem[Kaplan and Oudeyer(2004)]{Kaplan2004Maximizing-Learning-Progress:}
F.~Kaplan and P.-Y. Oudeyer.
\newblock Maximizing learning progress: An internal reward system for
  development.
\newblock {\em Embodied Artificial Intelligence}, pages 259--270, 2004.

\bibitem[Klyubin et~al.(2004)Klyubin, Polani, and
  Nehaniv]{Klyubin2004Organization-of-the-information-flow}
A.~S. Klyubin, D.~Polani, and C.~L. Nehaniv.
\newblock Organization of the information flow in the perception-action loop of
  evolved agents.
\newblock In {\em Evolvable Hardware, 2004. Proceedings. 2004 NASA/DoD
  Conference on}, pages 177--180, 2004.

\bibitem[Little and
  Sommer(2013)]{Little2013Learning-and-exploration-in-action-perception}
D.~Y. Little and F.~T. Sommer.
\newblock Learning and exploration in action-perception loops.
\newblock {\em Frontiers in Neural Circuits}, 7\penalty0 (37), 2013.

\bibitem[Markeli{\'c} and Zahedi(2007)]{Markelic2007An-Evolved-Neural}
I.~Markeli{\'c} and K.~Zahedi.
\newblock An evolved neural network for fast quadrupedal locomotion.
\newblock In Ming Xie and Steven Dubowsky, editors, {\em Advances in Climbing
  and Walking Robots, Proceedings of 10th International Conference {(CLAWAR
  2007)}}, pages 65--72. World Scientific Publishing Company, 2007.

\bibitem[Martius et~al.(2013)Martius, Der, and Ay]{MartiusDerAy2013}
G.~Martius, R.~Der, and N.~Ay.
\newblock Information driven self-organization of complex robotic behaviors.
\newblock {\em PLoS ONE}, 8\penalty0 (5):\penalty0 e63400, 05 2013.

\bibitem[Martius and Herrmann(2012)]{martiusherrmann:variantsofgso12}
G.~Martius and J.~M. Herrmann.
\newblock Variants of guided self-organization for robot control.
\newblock {\em Theory in Biosci.}, 131\penalty0 (3):\penalty0 129--137, 2012.
\newblock ISSN 1431-7613.

\bibitem[Martius et~al.(2012)Martius, Hesse, G{\"u}ttler, and Der]{lpzrobots12}
G.~Martius, F.~Hesse, F.~G{\"u}ttler, and R.~Der.
\newblock \textsc{LpzRobots}: A free and powerful robot simulator, version 0.7.
\newblock \url{http://robot.informatik.uni-leipzig.de/software}, 2012.

\bibitem[Oudeyer et~al.(2007)Oudeyer, Kaplan, and
  Hafner]{Oudeyer2007Intrinsic-Motivation-Systems}
P.-Y. Oudeyer, F.~Kaplan, and V.~V. Hafner.
\newblock Intrinsic motivation systems for autonomous mental development.
\newblock {\em IEEE Trans.~on Evo.~Computation}, 11\penalty0 (2):\penalty0
  265--286, 2007.

\bibitem[Pasemann et~al.(1999)Pasemann, Steinmetz, and
  Dieckman]{Pasemann1999Evolving-structure-and-function}
F.~Pasemann, U.~Steinmetz, and U.~Dieckman.
\newblock Evolving structure and function of neurocontrollers.
\newblock In {\em Proc. Congress Evolutionary Computation CEC 99}, volume~3,
  1999.

\bibitem[Pfeifer and Bongard(2006)]{Pfeifer2006How-the-Body}
R.~Pfeifer and J.~C. Bongard.
\newblock {\em How the Body Shapes the Way We Think: A New View of
  Intelligence}.
\newblock The MIT Press (Bradford Books), 2006.

\bibitem[Pfeifer et~al.(2007)Pfeifer, Lungarella, and
  Iida]{Pfeifer2007Self-Organization-Embodiment-and}
R.~Pfeifer, M.~Lungarella, and F.~Iida.
\newblock Self-organization, embodiment, and biologically inspired robotics.
\newblock {\em Science}, 318\penalty0 (5853):\penalty0 1088--1093, 2007.

\bibitem[Prokopenko et~al.(2006)Prokopenko, Gerasimov, and
  Tanev]{Prokopenko2006Evolving-Spatiotemporal-Coordination}
M.~Prokopenko, V.~Gerasimov, and I.~Tanev.
\newblock Evolving spatiotemporal coordination in a modular robotic system.
\newblock In {\em Proc.~SAB'06}, volume 4095, pages 558--569, 2006.

\bibitem[Schmidhuber(1990)]{Schmidhuber1990A-possibility-for}
J.~Schmidhuber.
\newblock A possibility for implementing curiosity and boredom in
  model-building neural controllers.
\newblock In {\em Proceedings of SAB'90}, pages 222--227, 1990.

\bibitem[Schmidhuber(1991)]{Schmidhuber1991Curious-Model-Building-Control}
J.~Schmidhuber.
\newblock Curious model-building control systems.
\newblock In {\em In Proc. International Joint Conference on Neural Networks,
  Singapore}, pages 1458--1463. IEEE, 1991.

\bibitem[Schmidhuber(2006)]{Schmidhuber:06cs}
J.~Schmidhuber.
\newblock Developmental robotics, optimal artificial curiosity, creativity,
  music, and the fine arts.
\newblock {\em Connection Science}, 18\penalty0 (2):\penalty0 173--187, 2006.

\bibitem[Sehnke et~al.(2010)Sehnke, Osendorfer, R{\"u}ckstiess, Graves, Peters,
  and Schmidhuber]{Sehnke2010Parameter-exploring-policy-gradients}
F.~Sehnke, C.~Osendorfer, T.~R{\"u}ckstiess, A.~Graves, J.~Peters, and
  J.~Schmidhuber.
\newblock Parameter-exploring policy gradients.
\newblock {\em Neural Netw}, 23\penalty0 (4):\penalty0 551--9, May 2010.

\bibitem[Storck et~al.(1995)Storck, Hochreiter, and Schmidhuber]{Storck:95}
J.~Storck, S.~Hochreiter, and J.~Schmidhuber.
\newblock Reinforcement driven information acquisition in non-deterministic
  environments.
\newblock In {\em Proceedings of the International Conference on Artificial
  Neural Networks, Paris}, volume~2, pages 159--164. EC2 \& Cie, 1995.

\bibitem[Sutton and Barto(1998)]{Sutton1998Reinforcement-Learning:-An}
R.~S. Sutton and A.~G. Barto.
\newblock {\em Reinforcement Learning: An Introduction}.
\newblock MIT Press, 1998.

\bibitem[von Twickel et~al.(2011)von Twickel, B{\"u}schges, and
  Pasemann]{Twickel2011Deriving-neural-network}
A.~von Twickel, A.~B{\"u}schges, and F.~Pasemann.
\newblock Deriving neural network controllers from neuro-biological data:
  implementation of a single-leg stick insect controller.
\newblock {\em Biological Cybernetics}, 104:\penalty0 95--119, 2011.

\bibitem[von Uexkuell(1934)]{Uexkuell1957A-Stroll-Through}
J.~von Uexkuell.
\newblock A stroll through the worlds of animals and men.
\newblock In C.~H. Schiller, editor, {\em Instinctive Behavior}, pages 5--80.
  International Universities Press, New York, 1934.

\bibitem[Yi et~al.(2011)Yi, Gomez, and Schmidhuber]{sunyi2011agi}
S.~Yi, F.~Gomez, and J.~Schmidhuber.
\newblock Planning to be surprised: Optimal {Bayesian} exploration in dynamic
  environments.
\newblock In {\em {Proc. Fourth Conference on Artificial General Intelligence
  (AGI), Google, Mountain View, CA}}, 2011.

\bibitem[Zahedi and Ay(2013)]{Zahedi2013Quantifying-Morphological-Computation}
K.~Zahedi and N.~Ay.
\newblock Quantifying morphological computation.
\newblock {\em Entropy}, 15\penalty0 (5):\penalty0 1887--1915, 2013.

\bibitem[Zahedi et~al.(2010)Zahedi, Ay, and
  Der]{Zahedi2010Higher-coordination-with}
K.~Zahedi, N.~Ay, and R.~Der.
\newblock Higher coordination with less control -- a result of information
  maximization in the sensori-motor loop.
\newblock {\em Adaptive Behavior}, 18\penalty0 (3--4):\penalty0 338--355, 2010.

\bibitem[Zahedi et~al.(2008)Zahedi, von Twickel, and
  Pasemann]{Zahedi2008YARS:-A-Physical}
K.~Zahedi, A.~von Twickel, and F.~Pasemann.
\newblock Yars: A physical 3d simulator for evolving controllers for real
  robots.
\newblock In S.~Carpin and et~al., editors, {\em SIMPAR 2008}, LNAI 5325, pages
  71---82. Springer, 2008.

\end{thebibliography}

\end{document}